\documentclass[journal,onecolumn]{IEEEtran}

\IEEEoverridecommandlockouts

\usepackage[utf8]{inputenc}
\usepackage{amsmath, scalerel} 
\usepackage{amssymb}
\usepackage{amsopn}
\usepackage{graphicx}
\usepackage[style=base]{caption}
\usepackage{color}
\usepackage{quotmark}
\usepackage{booktabs}
\usepackage{todonotes}
\usepackage{lipsum}
\usepackage{import}
\usepackage{footnote}
\usepackage{multirow}
\usepackage{hhline}
\usepackage{subcaption}
\usepackage[pdfborder={0 0 0}, bookmarks=false, final]{hyperref}

\usepackage{floatrow}

\usepackage[abbreviations,binary-units]{siunitx}
\DeclareSIUnit{\mAh}{mAh}
\DeclareSIUnit{\Wh}{Wh}

\usepackage{todonotes}
\usepackage[gernotes]{optional}

\usepackage{scrextend}
\addtokomafont{labelinglabel}{\sffamily}

\usepackage{tikz}
\usetikzlibrary{calc}
\usetikzlibrary{matrix}
\usetikzlibrary{math}
\usepackage{icomma}
\usetikzlibrary{plotmarks}
\usetikzlibrary{positioning}
\usetikzlibrary{shapes}
\usetikzlibrary{arrows}
\usetikzlibrary{arrows.meta}

\usepackage{pgfplots}
\pgfplotsset{compat=newest}
\usepgfplotslibrary{patchplots}
\usepackage{grffile}
\usepackage{ textcomp }

\usepackage{pgfplots}
\pgfplotsset{compat=newest} 
\pgfplotsset{plot coordinates/math parser=false} 
\newlength\figureheight 
\newlength\figurewidth 

\title{\LARGE \bf
M²VAE -- Derivation of a Multi-Modal Variational Autoencoder Objective from the Marginal Joint Log-Likelihood
}

\author{Timo Korthals$^{\dagger *}$
\thanks{
	$^{\dagger}$Bielefeld University,
	Cognitronics \& Sensor Systems,
	Inspiration 1, 33619 Bielefeld, Germany\newline
	\indent{$^{*}${\tt\small tkorthals@cit-ec.uni-bielefeld.de}}\newline
	\indent{}\textbf{Work in progress}\newline
	}
}

\makeatletter
\newcommand{\myspecialcell}[1]{\ifmeasuring@#1\else\omit$\displaystyle#1$\ignorespaces\fi}
\makeatother

\renewcommand{\a}{a}
\renewcommand{\b}{b}
\renewcommand{\c}{c}
\newcommand{\A}{A}
\newcommand{\B}{B}
\newcommand{\C}{C}
\newcommand{\texta}{\text{a}}
\newcommand{\textb}{\text{b}}
\newcommand{\textc}{\text{c}}
\DeclareMathOperator\dkl{D_{KL}}
\DeclareMathOperator\VarInfo{VI}
\DeclareMathOperator\Info{I}
\DeclareMathOperator\Ent{H}
\newcommand{\p}{p}
\newcommand{\q}{q}
\newcommand{\elbo}{\mathcal{L}}
\newcommand{\elboJ}{\mathcal{L}_{\text{J}}}
\newcommand{\elboTJ}{\mathcal{L}_{\widetilde{\text{J}}}}
\newcommand{\elboM}{\mathcal{L}_{\text{M}}}
\newcommand{\elboMa}{\elbo_{\text{M}_\texta}}
\newcommand{\elboMb}{\elbo_{\text{M}_\textb}}

\newcommand{\elboTM}{\elbo_{\widetilde{\text{M}}}}
\newcommand{\elboTMc}{\elbo_{\widetilde{\text{M}}_\textc}}
\newcommand{\elboTMa}{\elbo_{\widetilde{\text{M}}_\texta}}
\newcommand{\elboTMb}{\elbo_{\widetilde{\text{M}}_\textb}}
\newcommand{\optext}[1]{\myspecialcell{\hfill\text{#1}}}
\newcommand{\taref}[1]{\optext{\autoref{#1}}}
\renewcommand{\(}{\mathopen{}\left(}
\renewcommand{\)}{\right)\mathclose{}}

\DeclareMathOperator{\EX}{\mathbb{E}}

\DeclareMathOperator{\diffdoperator}{d}
\newcommand{\diffd}{\diffdoperator\!}

\begin{document}

\maketitle
\thispagestyle{empty}
\pagestyle{empty}

\begin{abstract}

This work gives an in-depth derivation of the trainable evidence lower bound (ELBO) obtained from the marginal joint log-Likelihood with the goal of training a multi-modal variational Autoencoder (M²VAE).

\end{abstract}

\section{Introduction}
\label{sec:intro}

\begin{figure}[h!]
	\footnotesize
\begingroup
  \makeatletter
  \providecommand\color[2][]{
    \errmessage{(Inkscape) Color is used for the text in Inkscape, but the package 'color.sty' is not loaded}
    \renewcommand\color[2][]{}
  }
  \providecommand\transparent[1]{
    \errmessage{(Inkscape) Transparency is used (non-zero) for the text in Inkscape, but the package 'transparent.sty' is not loaded}
    \renewcommand\transparent[1]{}
  }
  \providecommand\rotatebox[2]{#2}
  \newcommand*\fsize{\dimexpr\f@size pt\relax}
  \newcommand*\lineheight[1]{\fontsize{\fsize}{#1\fsize}\selectfont}
  \ifx\svgwidth\undefined
    \setlength{\unitlength}{237.45672271bp}
    \ifx\svgscale\undefined
      \relax
    \else
      \setlength{\unitlength}{\unitlength * \real{\svgscale}}
    \fi
  \else
    \setlength{\unitlength}{\svgwidth}
  \fi
  \global\let\svgwidth\undefined
  \global\let\svgscale\undefined
  \makeatother
  \begin{picture}(1,0.3730596)
    \lineheight{1}
    \setlength\tabcolsep{0pt}
    \put(0,0){\includegraphics[width=\unitlength,page=1]{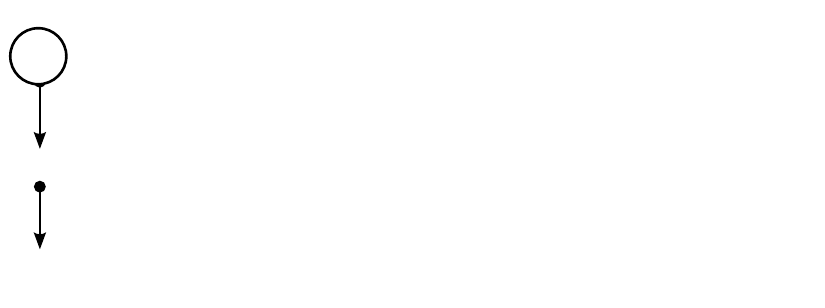}}
    \put(0.03706168,0.3139695){\color[rgb]{0,0,0}\makebox(0,0)[lt]{\begin{minipage}{0.58836782\unitlength}\raggedright $a$\end{minipage}}}
    \put(0,0){\includegraphics[width=\unitlength,page=2]{VAEs_trimodal.pdf}}
    \put(0.03830642,0.1618877){\color[rgb]{0,0,0}\makebox(0,0)[lt]{\begin{minipage}{0.58836782\unitlength}\raggedright $z$\end{minipage}}}
    \put(0,0){\includegraphics[width=\unitlength,page=3]{VAEs_trimodal.pdf}}
    \put(0.03706169,0.04467958){\color[rgb]{0,0,0}\makebox(0,0)[lt]{\begin{minipage}{0.58836782\unitlength}\raggedright $a'$\end{minipage}}}
    \put(0,0){\includegraphics[width=\unitlength,page=4]{VAEs_trimodal.pdf}}
    \put(0.26759832,0.16271529){\color[rgb]{0,0,0}\makebox(0,0)[lt]{\begin{minipage}{0.58836782\unitlength}\raggedright $z$\end{minipage}}}
    \put(0,0){\includegraphics[width=\unitlength,page=5]{VAEs_trimodal.pdf}}
    \put(0.16830635,0.31479721){\color[rgb]{0,0,0}\makebox(0,0)[lt]{\begin{minipage}{0.58836782\unitlength}\raggedright $a$\end{minipage}}}
    \put(0,0){\includegraphics[width=\unitlength,page=6]{VAEs_trimodal.pdf}}
    \put(0.3721535,0.32111408){\color[rgb]{0,0,0}\makebox(0,0)[lt]{\begin{minipage}{0.58836782\unitlength}\raggedright $b$\end{minipage}}}
    \put(0,0){\includegraphics[width=\unitlength,page=7]{VAEs_trimodal.pdf}}
    \put(0.19316021,0.04550729){\color[rgb]{0,0,0}\makebox(0,0)[lt]{\begin{minipage}{0.58836782\unitlength}\raggedright $a'$\end{minipage}}}
    \put(0,0){\includegraphics[width=\unitlength,page=8]{VAEs_trimodal.pdf}}
    \put(0.33383797,0.04467966){\color[rgb]{0,0,0}\makebox(0,0)[lt]{\begin{minipage}{0.58836782\unitlength}\raggedright $b'$\end{minipage}}}
    \put(0,0){\includegraphics[width=\unitlength,page=9]{VAEs_trimodal.pdf}}
    \put(0.10693008,0.23191506){\color[rgb]{0,0,0}\makebox(0,0)[lt]{\lineheight{1.25}\smash{\begin{tabular}[t]{l}$q_{\phi_{\texta}}$\end{tabular}}}}
    \put(0.35965025,0.20033078){\color[rgb]{0,0,0}\makebox(0,0)[lt]{\lineheight{1.25}\smash{\begin{tabular}[t]{l}$q_{\phi_{\textb}}$\end{tabular}}}}
    \put(0.11851867,0.0908136){\color[rgb]{0,0,0}\makebox(0,0)[lt]{\lineheight{1.25}\smash{\begin{tabular}[t]{l}$p_{\theta_{\texta}}$\end{tabular}}}}
    \put(0.35860498,0.11608102){\color[rgb]{0,0,0}\makebox(0,0)[lt]{\lineheight{1.25}\smash{\begin{tabular}[t]{l}$p_{\theta_{\textb}}$\end{tabular}}}}
    \put(0.21604069,0.31629305){\color[rgb]{0,0,0}\makebox(0,0)[lt]{\lineheight{1.25}\smash{\begin{tabular}[t]{l}$q_{\phi_{\texta\textb}}$\end{tabular}}}}
    \put(0.49758374,0.10502948){\color[rgb]{0,0,0}\makebox(0,0)[lt]{\lineheight{1.25}\smash{\begin{tabular}[t]{l}$p_{\theta_{\texta}}$\end{tabular}}}}
    \put(0.76981398,0.35450694){\color[rgb]{0,0,0}\makebox(0,0)[lt]{\lineheight{1.25}\smash{\begin{tabular}[t]{l}$q_{\phi_{\texta\textb\textc}}$\end{tabular}}}}
    \put(0,0){\includegraphics[width=\unitlength,page=10]{VAEs_trimodal.pdf}}
    \put(0.64283063,0.16271529){\color[rgb]{0,0,0}\makebox(0,0)[lt]{\begin{minipage}{0.58836782\unitlength}\raggedright $z$\end{minipage}}}
    \put(0,0){\includegraphics[width=\unitlength,page=11]{VAEs_trimodal.pdf}}
    \put(0.5026784,0.31479721){\color[rgb]{0,0,0}\makebox(0,0)[lt]{\begin{minipage}{0.58836782\unitlength}\raggedright $a$\end{minipage}}}
    \put(0,0){\includegraphics[width=\unitlength,page=12]{VAEs_trimodal.pdf}}
    \put(0.78822548,0.31479721){\color[rgb]{0,0,0}\makebox(0,0)[lt]{\begin{minipage}{0.58836782\unitlength}\raggedright $c$\end{minipage}}}
    \put(0,0){\includegraphics[width=\unitlength,page=13]{VAEs_trimodal.pdf}}
    \put(0.50267848,0.05182427){\color[rgb]{0,0,0}\makebox(0,0)[lt]{\begin{minipage}{0.58836782\unitlength}\raggedright $a'$\end{minipage}}}
    \put(0,0){\includegraphics[width=\unitlength,page=14]{VAEs_trimodal.pdf}}
    \put(0.78224103,0.05099664){\color[rgb]{0,0,0}\makebox(0,0)[lt]{\begin{minipage}{0.58836782\unitlength}\raggedright $c'$\end{minipage}}}
    \put(0,0){\includegraphics[width=\unitlength,page=15]{VAEs_trimodal.pdf}}
    \put(0.64041803,0.05182427){\color[rgb]{0,0,0}\makebox(0,0)[lt]{\begin{minipage}{0.58836782\unitlength}\raggedright $b'$\end{minipage}}}
    \put(0,0){\includegraphics[width=\unitlength,page=16]{VAEs_trimodal.pdf}}
    \put(0.64673498,0.32111408){\color[rgb]{0,0,0}\makebox(0,0)[lt]{\begin{minipage}{0.58836782\unitlength}\raggedright $b$\end{minipage}}}
    \put(0.47870178,0.35450694){\color[rgb]{0,0,0}\makebox(0,0)[lt]{\lineheight{1.25}\smash{\begin{tabular}[t]{l}$q_{\phi_{\texta\textc}}$\end{tabular}}}}
    \put(0.47536051,0.18386058){\color[rgb]{0,0,0}\makebox(0,0)[lt]{\lineheight{1.25}\smash{\begin{tabular}[t]{l}$q_{\phi_{\texta}}$\end{tabular}}}}
    \put(0.78729207,0.18386058){\color[rgb]{0,0,0}\makebox(0,0)[lt]{\lineheight{1.25}\smash{\begin{tabular}[t]{l}$q_{\phi_{\textc}}$\end{tabular}}}}
    \put(0.59728009,0.08497016){\color[rgb]{0,0,0}\makebox(0,0)[lt]{\lineheight{1.25}\smash{\begin{tabular}[t]{l}$p_{\theta_{\textb}}$\end{tabular}}}}
    \put(0.76751558,0.10603278){\color[rgb]{0,0,0}\makebox(0,0)[lt]{\lineheight{1.25}\smash{\begin{tabular}[t]{l}$p_{\theta_{\textc}}$\end{tabular}}}}
    \put(0,0){\includegraphics[width=\unitlength,page=17]{VAEs_trimodal.pdf}}
    \put(0.62764814,0.24675354){\color[rgb]{0,0,0}\makebox(0,0)[lt]{\lineheight{1.25}\smash{\begin{tabular}[t]{l}$q_{\phi_{\textb}}$\end{tabular}}}}
    \put(0.53395448,0.2124725){\color[rgb]{0,0,0}\makebox(0,0)[lt]{\lineheight{1.25}\smash{\begin{tabular}[t]{l}$q_{\phi_{\texta\textb}}$\end{tabular}}}}
    \put(0.71876032,0.2124725){\color[rgb]{0,0,0}\makebox(0,0)[lt]{\lineheight{1.25}\smash{\begin{tabular}[t]{l}$q_{\phi_{\textb\textc}}$\end{tabular}}}}
    \put(0,0){\includegraphics[width=\unitlength,page=18]{VAEs_trimodal.pdf}}
  \end{picture}
\endgroup

	\caption{Evolution of full uni-, bi-, and tri-modal VAEs comprising all modality permutations}
	\label{fig:VAE}
\end{figure}

\textit{Variational auto encoder} (VAE) combine neural networks with variational inference to allow unsupervised learning of complicated distributions according to the graphical model shown in \autoref{fig:VAE} (left).
A $D_\a$-dimensional observation $\a$ is modeled in terms of a $D_z$-dimensional latent vector $z$ using a probabilistic decoder $\p_{\theta_\texta}\(z\)$ with parameters $\theta$.
To generate the corresponding embedding $z$ from observation $\a$, a probabilistic encoder network with $\q_{\phi_\texta}\(z\)$ is being provided which parametrizes the posterior distribution from which $z$ is sampled.
The encoder and decoder, given by neural networks, are trained jointly to bring $\a$ close to an $\a'$ under the constraint that an approximate distribution needs to be close to a prior $\p\(z\)$ and hence inference is basically learned during training.

The specific objective of VAEs is the maximization of the marginal distribution $\p\(\a\) = \int \p_{\theta}\(\a|z\)\p\(z\)\diffd\a$.
Because this distribution is intractable, the model is instead trained via \textit{stochastic gradient variational Bayes} (SGVB) by maximizing the \textit{evidence lower bound} (ELBO) $\elbo$ of the marginal log-likelihood as described in Sec. \ref{sec:vae}
This approach proposed by \cite{DBLP:journals/corr/KingmaW13} is used in settings where only a single modality $\a$ is present in order to find a latent encoding $z$ (c.f. \autoref{fig:VAE} (left)).

This work gives an in-depth derivation of the trainable \textit{evidence lower bound} (ELBO) obtained from the marginal joint log-Likelihood, that satisfies all plate models as depicted in Fig. \ref{fig:VAE}, we are with the goal of training a multi-modal variational Autoencoder (M²VAE).

\newpage
\section{Variational Autoencoder}
\label{sec:vae}
First, the derivation of the vanilla \textit{Variational Autoencoder} by \cite{DBLP:journals/corr/KingmaW13} is recaped.

\subsection{The Variational Bound}

\begin{align}
L &= \log\( \p\( \a \) \) &\\
  &= \sum_z \q \( z | \a \)  \log \( \p \( \a \) \)&\optext{\autoref{eq:mariginal_likelihood} w/o conditional}\\
  &= \sum_z \q \( z | \a \)  \log \( \frac{\p \( z,\a \)}{ \p \( z|\a \) }   \) & \taref{eq:bayes}\\
  &= \sum_z \q \( z | \a \)  \log \( \frac{\p \( z,\a \)}{ \p \( z|\a \) } \frac{\q \( z|\a \)}{ \q \( z|\a \) }  \) & \optext{multiplied by 1} \\
  &= \sum_z \q \( z | \a \)  \log \( \frac{\p \( z,\a \)}{ \q \( z|\a \) } \frac{\q \( z|\a \)}{ \p \( z|\a \) }  \) & \optext{reordered} \\
  &= \sum_z \q \( z | \a \)  \log \( \frac{\p \( z,\a \)}{ \q \( z|\a \) }  \)
  +
  \sum_z \q \( z | \a \)  \log \( \frac{\q \( z|\a \)}{ \p \( z|\a \) }  \)& \taref{eq:logadd} \\
  &= \elbo +  \dkl \( \q \( z | \a \) \| \p\( z| \a \) \)  & \optext{\autoref{eq:kld} \& \ref{eq:elbo}} \\
  &\geq \elbo & \optext{$\dkl \geq 0$}
  \label{eq:loglikelihood_J}
\end{align}

$\dkl$ is the Kulbeck-Leibler divergence, with $\dkl\geq 0$, thats depends on how good $\q\(z|\a\)$ can approximate $\p\(z|\a\)$.
$\elbo$ is the lower variational bound of the marginal log-likelihood, also called the \textit{evidence lower bound} (\textbf{ELBO}).
If and only if the two distributions $\q$ and $\p$ are identical, $\dkl$ becomes 0 ($\q = \p \Leftrightarrow \dkl = 0$).
$\elbo = L$ means on the other hand therefore implicitly, that $q$ perfectly approximates $p$.
It is beacuse $\elbo$ and $\dkl$ are in equilibrium so that minimizing $\dkl$ is identical to the maximization of $\elbo$ ($\min \dkl \Leftrightarrow \max \elbo$).
Minimizing $\dkl$ is not feasible, because we don't know the true posterior $p\(z|\a\)$.
Therefore, $\elbo$ is further investigated.

\subsection{Approximate Inference (i.e. rewriting $\elbo$)}

\begin{align}
\elbo &= \sum_z \q \( z | \a \)  \log \( \frac{\p \( z,\a \)}{ \q \( z|\a \) }  \) & \\
 &=  \sum_z \q \( z | \a \)  \log \( \frac{\p \( \a | z \) \p\( z \)}{ \q \( z|\a \) }  \) &  \taref{eq:bayes}\\
 &=  \sum_z \q \( z | \a \)  \log \( \frac{\p\( z \)}{ \q \( z|\a \) }  \) 
 +
 \sum_z \q \( z | \a \)  \log \( \p \( \a | z \)  \)&  \taref{eq:logadd}\\
 &=  - \dkl \( \q \( z| \a \) \| \p\( z \) \) + \EX_{\q\( z | \a \)} \log \( \p \( \a | z \)  \) & \optext{\autoref{eq:kld}}
 \label{eq:vae_loss}
\end{align}

If the variable $\a$ is replaced by some real valued sample $\a^{\(i\)}$ (e.g. image or LiDAR scan), two terms can be identified:

\begin{align}
\elbo &= \underbrace{- \dkl \( \q_\phi \( z| \a^{\(i\)} \) \| \p\( z \) \)}_\text{Regularization}
+
\underbrace{\EX_{\q_\phi\( z | \a^{\(i\)} \)} \log \( \p_\theta \( \a^{\(i\)} | z \)  \)}_{\text{Reconstruction}}
\label{eq:reg_rec}
\end{align}

The first term is just a regularize that punishes the variational distribution $\q$, that is the approximatior of the posterior distribution, if it deviates from some prior $\p\( z \)$.
The reconstruction term on the other hand compares the difference between the data $\a^{\(i\)}$ of $\q_{\phi}\( z | \a^{\(i\)} \)$ wrt. the sampled data $\a^{\(i\)}$ from the likelihood function $\p_{\theta}\( \a^{\(i\)} | z \)$.

That means if $\elbo$ is going to be maximized, the posterior function has to be equal to some prior and the data that is used to sample the latent feature $z$ from the posterior should be equal to the data from the likelihood function $\p_{\theta}$.
The objective is now, to find a function $q_\phi$ and $p_\theta$ which own these properties.
Luckily, if some parametrized functions $\q_\phi$ and $\p_\theta$ (with parameters $\phi$ and $\theta$) are applied, every increase of \autoref{eq:reg_rec} means that the variational approximator $q_\phi\( z | \a \)$ comes closer to the real posterior functions $\p \( z | \a \)$.
Therefore, numerical optimization techniques like gradient descent can be applied to this issue.
Commonly, two neuronal network, where each tries to find the optimal parameters $\phi$ and $\theta$, are applied to approximate the functions $q_\phi$ (i.e. the encoder) and $\p_\theta$ (i.e. the decoder).
However, since the true value of $L$ remains unknown, maximization of $\elbo$ can only be done until convergence.
Thus, the overall procedure has the property of finding local optima.

\newpage

\section{Joint Variational Autoencoder}
\label{sec:joint_VAE}

Second, we expand the VAE from Sec. \ref{sec:vae} to the marginal joint log-likelihood and derive the variational bound as follows:

\begin{align}
L_{\text{J}} &= \log\( \p\( \a,\b \) \) &\\
&= \sum_z \q \( z | \a,\b \)  \log \( \p \( \a,\b \) \)&\optext{\autoref{eq:mariginal_likelihood} w/o conditional}\\
&= \sum_z \q \( z | \a,\b \)  \log \( \frac{\p \( z,\a,\b\)}{ \p \( z|\a,\b \) }   \) & \taref{eq:bayes_multi}\\
&= \sum_z \q \( z | \a,\b \)  \log \( \frac{\p \( z,\a,\b \)}{ \p \( z|\a,\b \) } \frac{\q \( z|\a,\b \)}{ \q \( z|\a,\b \) }  \) & \optext{multiplied by 1} \\
&= \sum_z \q \( z | \a,\b \)  \log \( \frac{\p \( z,\a,\b \)}{ \q \( z|\a,\b \) } \frac{\q \( z|\a,\b \)}{ \p \( z|\a,\b \) }  \) & \optext{reordered} \\
&= \sum_z \q \( z | \a,\b \)  \log \( \frac{\p \( z,\a,\b \)}{ \q \( z|\a,\b \) }  \)
+
\sum_z \q \( z | \a,\b \)  \log \( \frac{\q \( z|\a,\b \)}{ \p \( z|\a,\b \) }  \)& \taref{eq:logadd} \\
&= \elbo_{\text{J}} +  \dkl \( \q \( z | \a,\b \) \| \p\( z| \a,\b \) \)  & \optext{\autoref{eq:kld} \& \ref{eq:elbo}} \\
&\geq \elboJ &
\end{align}

Approximate Inference (i.e. rewriting $\elboJ$):

\begin{align}
\elboJ &= \sum_z \q \( z | \a,\b \)  \log \( \frac{\p \( z,\a,\b \)}{ \q \( z|\a,\b \) }  \) & \\
&=  \sum_z \q \( z | \a,\b \)  \log \( \frac{\p \( \a,\b | z \) \p\( z \)}{ \q \( z|\a,\b \) }  \) &  \taref{eq:bayes}\\
&=  \sum_z \q \( z | \a,\b \)  \log \( \frac{\p\( z \)}{ \q \( z|\a,\b \) }  \) 
+
\sum_z \q \( z | \a,\b \)  \log \( \p \( \a,\b | z \)  \)&  \taref{eq:logadd}\\
&=  \sum_z \q \( z | \a,\b \)  \log \( \frac{\p\( z \)}{ \q \( z|\a,\b \) }  \) 
+
\sum_z \q \( z | \a,\b \)  \log \( \p \( \a | z \)  \)
+
\sum_z \q \( z | \a,\b \)  \log \( \p \( \b | z \)  \)&  \taref{eq:iid}\\
&=  - \dkl \( \q \( z| \a,\b \) \| \p\( z \) \) + \EX_{\q\( z | \a,\b \)} \log \( \p \( \a | z \)  \) + \EX_{\q\( z | \a,\b \)} \log \( \p \( \b | z \)  \) & \optext{\autoref{eq:kld}}
\label{eq:joint_vae_loss}
\end{align}

Three different terms can be identified:

\begin{align}
\elboJ =
	\underbrace{- \dkl \( \q_{\phi_{\texta\textb}} \( z| \a,\b \) \| \p\( z \) \)}_{\text{Regularization}}
	+
	\underbrace{\EX_{\q_{\phi_{\texta\textb}}\( z | \a,\b \)} \log \( \p_{\theta_\texta} \( \a | z \)  \)}_{\text{Reconstruction wrt. $\a$}}
	+
	\underbrace{\EX_{\q_{\phi_{\texta\textb}}\( z | \a,\b \)} \log \( \p_{\theta_\textb} \( \b | z \)  \)}_{\text{Reconstruction wrt. $\b$}}
\label{eq:joint_reg_rec_rec}
\end{align}

A regularization for the joint encoder $\q_{\phi_{\texta\textb}}$ and two reconstruction terms, one for each decoder $\p_{\theta_{\texta}}$ and $\p_{\theta_{\textb}}$.

\newpage

\section{Joint Multi-Modal Variational Autoencoder via Variation of Information}

The issue with the joint VAE is the lacking possibility of encoding just one modality $\a$ or $\b$.
Thus, Suzuki et al. \cite{Suzuki2017} exploit the \textit{Variation of Information} (VI) and derive the evidence lower bound wrt. the VI.

First, the conditional probability is investigated

\begin{align}
\p\( \a | \b\) &= \frac{\p\( z,\a | \b\)}{\p\( z | \a,\b\)} & \taref{eq:bayes_multi}\\
&= \frac{1}{\p\( z | \a,\b\)}\p\( z,\a | \b\) &\\
&= \frac{1}{\p\( z | \a,\b\)}\frac{\p\( z,\a,\b\)}{\p\( \b \)} & \taref{eq:bayes}\\
&= \frac{1}{\p\( z | \a,\b\)}\frac{\p\( \a,\b|z\)\p\( z \)}{\p\( \b \)} & \taref{eq:bayes}\\
&= \frac{1}{\p\( z | \a,\b\)}\frac{\p\( \a|z\) \p\( \b|z\) \p\( z \)}{\p\( \b \)} & \taref{eq:iid}\\
&= \frac{1}{\p\( z | \a,\b\)}\frac{\p\( \a | z\) \p\( z|\b\) \frac{\p\( \b \)}{\p\( z \)} \p\( z \)}{\p\( \b \)} & \taref{eq:bayes}\\
&= \frac{p\( \a | z \) \p\( z | \b \)}{\p\( z | \a,\b\)} & 
\label{eq:jmvae_conditional}
\end{align}

Further, the marginal log-likelihood of a conditional distribution can be written as:

\begin{align}
L_{\text{M}_{\texta}} &= \log\(\p\( \a | \b\)\) & \\
 &= \sum_z \q \( z | \a,\b \)  \log \( \p \( \a|\b \) \)&\optext{\autoref{eq:mariginal_likelihood} w/o conditional}\\
&= \sum_z \q \( z | \a,\b \)  \log \( \frac{\p \( z,\a|\b\)}{ \p \( z|\a,\b \) }   \) & \taref{eq:bayes_multi}\\
&= \sum_z \q \( z | \a,\b \)  \log \( \frac{\p \( z,\a|\b \)}{ \p \( z|\a,\b \) } \frac{\q \( z|\a,\b \)}{ \q \( z|\a,\b \) }  \) & \optext{multiplied by 1} \\
&= \sum_z \q \( z | \a,\b \)  \log \( \frac{\p \( z,\a|\b \)}{ \q \( z|\a,\b \) } \frac{\q \( z|\a,\b \)}{ \p \( z|\a,\b \) }  \) & \optext{reordered} \\
&= \sum_z \q \( z | \a,\b \)  \log \( \frac{\p \( z,\a|\b \)}{ \q \( z|\a,\b \) }  \)
+
\sum_z \q \( z | \a,\b \)  \log \( \frac{\q \( z|\a,\b \)}{ \p \( z|\a,\b \) }  \)& \taref{eq:logadd} \\
&= \elboMa +  \dkl \( \q \( z | \a,\b \) \| \p\( z| \a,\b \) \)  & \optext{\autoref{eq:kld} \& \ref{eq:elbo}} \label{eq:elbo_conditional}\\
&\geq \elboMa &
\label{eq:elbo_conditional_geq}
\end{align}

Further, the log-likelihood of the VI can be written as

\begin{align}
L_\text{M} &= L_{\text{M}_{\texta}} + L_{\text{M}_{\textb}} & \\
 &= \log\(\p\( \a | \b\)\) + \log\(\p\( \b | \a\)\) & \\
 &= \elboMa + \elboMb + 2 \dkl \( \q \( z | \a,\b \) \| \p\( z| \a,\b \) \) & \taref{eq:elbo_conditional}\\
 &\geq \elboMa + \elboMb & 
 \label{eq:elbo_joint_geq}
\end{align}

\begin{align}
\elboMa + \elboMb &= \sum_z \q \( z | \a,\b \)  \log \( \frac{\p \( z,\a|\b \)}{ \q \( z|\a,\b \) }  \) + \sum_z \q \( z | \a,\b \)  \log \( \frac{\p \( z,\b|\a \)}{ \q \( z|\a,\b \) }  \) & \\
 &= \sum_z \q \( z | \a,\b \)  \log \( \frac{\p \( \a|z \) \p \( z|\b \)}{ \q \( z|\a,\b \) }  \)
 +
\sum_z \q \( z | \a,\b \)  \log \( \frac{\p \( \b|z \) \p \( z|\a \)}{ \q \( z|\a,\b \) }  \) & \taref{eq:jmvae_conditional} \\
 &= \sum_z \q \( z | \a,\b \)  \log \( \frac{\p \( \a|z \)}{ \q \( z|\a,\b \) }  \)
 +
 \sum_z \q \( z | \a,\b \)  \log \( \frac{\p \( z|\b \)}{ \q \( z|\a,\b \) }  \) & \\
 &\phantom{=} +
 \sum_z \q \( z | \a,\b \)  \log \( \frac{\p \( \b|z \)}{ \q \( z|\a,\b \) }  \)
 +
  \sum_z \q \( z | \a,\b \)  \log \( \frac{\p \( z|\a \)}{ \q \( z|\a,\b \) }  \) & \optext{reordering} \\
 &= \EX_{\q\( z | \a,\b \)} \log \( \p \( \a | z \)  \) - \dkl \( \q \( z | \a,\b \) \| \p\( z| \b \) \) & \\
 &\phantom{=} +
 \EX_{\q\( z | \a,\b \)} \log \( \p \( \b | z \)  \) - \dkl \( \q \( z | \a,\b \) \| \p\( z| \a \) \)& \optext{\autoref{eq:kld}}\\
 &= \EX_{\q\( z | \a,\b \)} \log \( \p \( \a | z \)  \) - \dkl \( \q \( z | \a,\b \) \| \p\( z| \b \) \) & \\
 &\phantom{=} +
 \EX_{\q\( z | \a,\b \)} \log \( \p \( \b | z \)  \) - \dkl \( \q \( z | \a,\b \) \| \p\( z| \a \) \)& \\
 &\phantom{=} + \dkl \( \q \( z | \a,\b \) \| \p\( z| \a,\b \) \) - \dkl \( \q \( z | \a,\b \) \| \p\( z| \a,\b \) \) & \optext{added 0}\\
 &= \elboJ - \dkl \( \q \( z | \a,\b \) \| \p\( z| \b \) \) - \dkl \( \q \( z | \a,\b \) \| \p\( z| \a \) \) + \dkl \( \q \( z | \a,\b \) \| \p\( z| \a,\b \) \)  & \optext{substitute eq. \ref{eq:joint_vae_loss}} \\
 &\geq \elboJ - \dkl \( \q \( z | \a,\b \) \| \p\( z| \b \) \) - \dkl \( \q \( z | \a,\b \) \| \p\( z| \a \) \) & \label{eq:multi_vae_loss}\\
 &=: \elboM
\end{align}

With respect to \autoref{eq:joint_reg_rec_rec}, the following regularization terms can be identified:

\begin{align}
\elboM = \elboJ - \underbrace{\dkl \( \q_{\phi_{\texta\textb}} \( z | \a,\b \) \| \q_{\phi_{\textb}}\( z| \b \) \)}_{\text{Unimodal PDF fitting of encoder b}} - \underbrace{\dkl \( \q_{\phi_{\texta\textb}} \( z | \a,\b \) \| \q_{\phi_{\texta}}\( z| \a \) \)}_{\text{Unimodal PDF fitting of encoder a}}
\label{eq:multi_reg_rec}
\end{align}

\subsection{Conclusion}
\label{sec:recap_conclusion}

The introduced KL regularization by Suzuki et al. \cite{Suzuki2017} tries to find a mean representative of all parameters between clusters in the latent space.
This is absolutely correct for the mean values but is insufficient for all other statistics of the distribution that are estimated by $\q_{\phi_{\text{*}}}$.
The approach would be sufficient, iff the true latent joint probability would be known, because only then the KL divergence is able to adapt to it in a correct manner.
Unfortunately, this is not the case and thus all the estimated statistic values of the uni-modal encoders $\q_{\phi_{\text{*}}}$, except the mean, are questionable.

\newpage

\section{Proposed Joint Multi-Modal Autoencoder}

\subsection{The Variational Bound}

\begin{align}
L_{\text{PJ}} &= \log\(\p\( \a,\b\)\) & \\
              &= \frac{2}{2}\log\(\p\( \a,\b\)\) & \optext{multiplied by 1}\\
              &= \frac{1}{2}\log\(\p\( \a,\b\)^2  \) & \taref{eq:logpower}\\
              &= \frac{1}{2}\log\(\p\( \a,\b\)\p\( \a,\b\)  \) & \\
              &= \frac{1}{2}\log\(\p\( \b \)\p\( \a|\b\)\p\( \b|\a\)\p\( \a \)  \) & \taref{eq:bayes}\\
              &= \frac{1}{2}\( \log\(\p\( \a \)\) +  \log\(\p\( \b|\a\)\) + \log\(\p\( \a|\b\) \) + \log\(\p\( \b \)\)  \) & \taref{eq:logadd}\\
              &\geq \frac{1}{2}\(\elbo_{\text{a}} + \elboMa + \elboMb + \elbo_{\text{b}}\) &  \optext{\autoref{eq:loglikelihood_J} \& \ref{eq:elbo_joint_geq}} \\
              &\geq \frac{1}{2}\(\elbo_{\text{a}} + \elboM + \elbo_{\text{b}}\) & \taref{eq:multi_reg_rec} \\
              &:= \elbo_{\text{PJ}}
\end{align}

\subsection{Approximating Inference (i.e. rewriting $\elbo_{\text{PJ}}$)}

\begin{align}
2\elbo_{\text{PJ}} &= \elbo_{\text{a}} + \elboM + \elbo_{\text{b}} & \\
                  &= - \dkl \( \q \( z| \a \) \| \p\( z \) \) + \EX_{\q\( z | \a \)} \log \( \p \( \a | z \)  \)& \optext{\autoref{eq:vae_loss}}\\
                  &\phantom{=} - \dkl \( \q \( z| \a,\b \) \| \p\( z \) \) + \EX_{\q\( z | \a,\b \)} \log \( \p \( \a | z \)  \) + \EX_{\q\( z | \a,\b \)} \log \( \p \( \b | z \)  \) & \taref{eq:joint_vae_loss} \\
                  &\phantom{=} - \dkl \( \q \( z | \a,\b \) \| \p\( z| \b \) \) - \dkl \( \q \( z | \a,\b \) \| \p\( z| \a \) \) & \optext{\autoref{eq:multi_vae_loss}} \\
                  &\phantom{=} - \dkl \( \q \( z| \b \) \| \p\( z \) \) + \EX_{\q\( z | \b \)} \log \( \p \( \b | z \)  \) & \optext{\autoref{eq:vae_loss}}
\end{align}

Applying the corresponding function approximators, the formula can be written as:

\begin{align}
2\elbo_{\text{PJ}} &= \elbo_{\text{a}} + \elboM + \elbo_{\text{b}} & \\
&= - \dkl \( \q_{\phi_{\texta}} \( z| \a \) \| \p\( z \) \) + \EX_{\q_{\phi_{\texta}}\( z | \a \)} \log \( \p_{\theta_{\texta}} \( \a | z \)  \)& \optext{\autoref{eq:reg_rec}}\label{eq:mm_reg_rec_Ua} \\
&\phantom{=} - \dkl \( \q_{\phi_{\texta\textb}} \( z| \a,\b \) \| \p\( z \) \) + \EX_{\q_{\phi_{\texta\textb}}\( z | \a,\b \)} \log \( \p_{\theta_{\texta}} \( \a | z \)  \) + \EX_{\q_{\phi_{\texta\textb}}\( z | \a,\b \)} \log \( \p_{\theta_{\textb}} \( \b | z \)  \) & \taref{eq:joint_reg_rec_rec} \label{eq:mm_reg_rec_J} \\
&\phantom{=} - \dkl \( \q_{\phi_{\texta\textb}} \( z | \a,\b \) \| \q_{\phi_{b}}\( z| \b \) \) - \dkl \( \q_{\phi_{\texta\textb}} \( z | \a,\b \) \| \q_{\phi_{\texta}}\( z| \a \) \) & \optext{\autoref{eq:multi_reg_rec}}\label{eq:mm_reg_rec_M} \\
&\phantom{=} - \dkl \( \q_{\phi_{\textb}} \( z| \b \) \| \p\( z \) \) + \EX_{\q_{\phi_{b}}\( z | \b \)} \log \( \p_{\theta_{\textb}} \( \b | z \)  \) & \optext{\autoref{eq:reg_rec}}\label{eq:mm_reg_rec_Ub}
\end{align}

Investigating every line of the formula, the following properties can be identified:
\autoref{eq:mm_reg_rec_J} is the common multi-modal VAE loss derived from the joint probability, while \autoref{eq:mm_reg_rec_M} adds the feature introduced by Suzuki et al. \cite{Suzuki2017}.
It introduces the KL regularization that brings the posterior distribution of an uni-modal encoder close to the distribution of the multi-modal case.
The drawback of this approach is discussed in \autoref{sec:recap_conclusion}.
The new lines, i.e. \autoref{eq:mm_reg_rec_Ua} and \ref{eq:mm_reg_rec_Ub}, introduce the regularization of the uni-modal encoders wrt. the common prior and the reconstruction loss.
The regularizer cares about the fact, that the uni-modal distribution does not deviate to much from the common prior while the reconstruction term shapes the remaining statistics including the mean.
However, the last fact is very important, while the mean value in latent space might not be the best representative of the likelihood (i.e. the decoded data).
This property cannot be respected by the KL divergence, but by the introduced reconstruction term.

\subsubsection{ of General Expression for Arbitrary Number of Modalities}

Comprising the applied steps to derive $\elbo_{\text{PJ}}$ from the former section, we can identify that by successively applying logarithm and Bayes rules, we derive the ELBO for the proposed multi-modal VAE as follows:
First, given the independent set of observable modalities $\mathcal{M}=\lbrace\texta,\textb,\textc,\ldots\rbrace$, its marginal log-likelihood $\log \p\(\mathcal{M}\)=:L_{\text{M\textsuperscript{2}}}$ is multiplied by the cardinality of the set as the neutral element $1=\frac{|\mathcal{M}|}{|\mathcal{M}|}$.
Second, applying logarithm multiplication rule, the nominator is written as the argument's exponent.
Third, Bayes rule is applied to each term wrt. the remaining observable modalities to derive their conditionals.
Therefore, we can write
\begin{align}
L_{\text{M\textsuperscript{2}}_{\mathcal{M}}} &= \log \p\(\mathcal{M} \) \overset{\text{mul. 1}}{=}
\frac{|\mathcal{M}|}{|\mathcal{M}|} \log \p\(\mathcal{M} \)
\overset{\text{log. mul.}}{=} 
\frac{1}{|\mathcal{M}|} \log \p\(\mathcal{M} \)^{|\mathcal{M}|}
\\
&\overset{\text{Bayes}}{=} 
\frac{1}{|\mathcal{M}|} \sum_{m\in\mathcal{M}} \log \p\(\mathcal{M} \setminus m  \) \p\(m | \mathcal{M} \setminus m  \)
\\
&\overset{\text{log. add}}{=} 
\frac{1}{|\mathcal{M}|} \sum_{m\in\mathcal{M}} \log \p\(\mathcal{M} \setminus m  \) + \log \p\(m | \mathcal{M} \setminus m  \)\textrm{.}
\end{align}

The expression $\sum_{m\in\mathcal{M}} \log \p\(m | \mathcal{M} \setminus m  \)$ is the general form of the marginal log-likelihood for the \textit{variation of information} (VI), as introduced by \cite{Suzuki2017} for the JMVAE, for any set $\mathcal{M}$.
Thus, it can be directly substituted with $L_{\text{M}_{\mathcal{M}}}$.
The expression $\sum_{m\in\mathcal{M}} \log \p\(\mathcal{M} \setminus m  \)$ is the combination of all joint log-likelihoods of the subsets of $\mathcal{M}$ which have one less element.
Therefore, this term can be rewritten as 
\begin{align}
\sum_{m\in\mathcal{M}} \log \p\(\mathcal{M} \setminus m  \) = \sum_{\widetilde{m}\in\widetilde{\mathcal{M}}} \log \p\( \widetilde{m} \)
\end{align}
with $\widetilde{\mathcal{M}}=\left\lbrace m | m \in \mathcal{P}\( \mathcal{M}\), |m|=|\mathcal{M}|-1 \right\rbrace$
Finally, $\log \p\( \widetilde{m} \)$ can be substituted by $L_{\text{M\textsuperscript{2}}_{\widetilde{m}}} $ without loss of generality.
However, it is worth noticing that substitution stops at the end of recursion and therefore, all final expressions $\log \p\( \widetilde{m} \)\ \forall \ |\widetilde{m}|\equiv1$ remain. $ \square $

This results in the final recursive log-likelihood expression from which the ELBO can be directly derived as follows:
\begin{align}
L_{\text{M\textsuperscript{2}}_{\mathcal{M}}} &= \frac{1}{|\mathcal{M}|} \( L_{\text{M}_{\mathcal{M}}} + \sum_{\widetilde{m}\in\widetilde{\mathcal{M}}} L_{\text{M\textsuperscript{2}}_{\widetilde{m}}} \) \geq \frac{1}{|\mathcal{M}|} \( \elbo_{\text{M}_{\mathcal{M}}} + \sum_{\widetilde{m}\in\widetilde{\mathcal{M}}} \elbo_{\text{M\textsuperscript{2}}_{\widetilde{m}}} \)
=: \elbo_{\text{M\textsuperscript{2}}_{\mathcal{M}}}\text{.} \label{eq:log_expression_induction}
\end{align}

\newpage

\section{Appendix}

Variants of Bayes equation:
\begin{align}
\p\( \a \) = \frac{\p \( z,\a \)}{ \p \( z|\a \)},\qquad
\p \( z|\a \) = \frac{\p \( z,\a \)}{ \p\( \a \)},\qquad
\p \( z,\a \) = \frac{\p \( \a \)}{ \p\( z,\a \)}
\label{eq:bayes}
\end{align}

\begin{align}
\p\( \a|\b,\c\)
\overset{\text{eq. \ref{eq:bayes}}}{=}
\frac{ \p\( \a,\b | \c \) }{ \p\( \b|\c \) } 
\overset{\text{eq. \ref{eq:bayes}}}{=}
\frac{ \p\( \a,\b,\c \) }{ \p\( \b|\c \) \p\( \c \)} 
\overset{\text{eq. \ref{eq:bayes}}}{=}
\frac{ \p\( \a,\b,\c \) }{ \p\( \b,\c \)}    
\label{eq:bayes_multi}
\end{align}

Logarithm rules:
\begin{align}
\log\( \a \b \) = \log\( \a \) + \log\( \b \)
\label{eq:logadd}
\end{align}
\begin{align}
\a\log\( \b \) = \log\( \b^\a \)
\label{eq:logpower}
\end{align}

Evidence lower bound:
\begin{align}
\elbo = \sum_z \q \( z | \a \)  \log \( \frac{\p \( z,\a \)}{ \q \( z|\a \) }  \)
\label{eq:elbo}
\end{align}

Marginal likelihood:
\begin{align}
\p \( \a | \b \) = \sum_z \p \( \a | z \)  \p \( z | \b \)
\label{eq:mariginal_likelihood}
\end{align}

Independent and identically distributed random variables (i.i.d. or iid or IID):
\begin{align}
\p\( \a,\b,\c \) = \p \( \a \) \p \( \b \) \p \( \c \)
\label{eq:iid}
\end{align}

\subsection{Kulbeck-Leibler Divergence}

\begin{align}
\dkl \( \q \( z | \a \) \| \p\( z| \a \) \) = \sum_z \q \( z | \a \)  \log \( \frac{\q \( z|\a \)}{ \p \( z|\a \) }  \)
\label{eq:kld}
\end{align}
\begin{align}
\dkl \( \mathcal{N}_1 \(\mu_1,\sigma_1 \) \| \mathcal{N}_2 \(\mu_2,\sigma_2 \) \)
=
\log \( \sigma_2 \) - \log\( \sigma_1 \) + \frac{\sigma_1^2}{2\sigma_2^2} + \frac{\( \mu_1 - \mu_2 \)^2}{2\sigma_2^2} - \frac{1}{2}
\label{eq:kld}
\end{align}

more tbd.

\subsection{Variation of Information}

\subsubsection{Operator Names}

\begin{itemize}
	\item $\VarInfo\( \A,\B \)$: Variation of Information between some properties $\A$ and $\B$
	\item $\Info\( \A\)$: Information of $\A$ (or mutual information)
	\item $\Info\( \A,\B\)$: \textit{Mutual Information} (MI) of $\A$ and $\B$
	\item $\Info\( \A,\B|\C\)$: \textit{Mutual Conditional Information} (MCI) of $\A$ and $\B$ given $\C$
	\item $\Ent\( \A \)$: Entropy of $\A$
	\item $\Ent\( \A , \B\)$: \textit{Joint Entropy} (JE) of $\A$ and $\B$
	\item $\Ent\( \A | \B\)$: \textit{Conditional Entropy} (CE) of $\A$ given $\B$
\end{itemize}

The \textit{Variation of Information} (VI) between some random variables can be written as 
\begin{align}
\VarInfo \( \A,\B \) = \Ent \( \A \) + \Ent \( \B \) - 2 \Info \( \A,\B \) = \Ent \( \A | \B \) + \Ent \( \B | \A \)
\end{align}
and
\begin{align}
\VarInfo \( \A,\B,\C \) = \Ent \( \A \) + \Ent \( \B \) + \Ent \( \C \)- 3 \Info \( \A,\B \) = \Ent \( \A | \B,\C\) + \Ent \( \B | \A,\C \) + \Ent \( \C | \A,\B\)
\end{align}

and so on \ldots.

\begin{figure*}
	\footnotesize
\begingroup
  \makeatletter
  \providecommand\color[2][]{
    \errmessage{(Inkscape) Color is used for the text in Inkscape, but the package 'color.sty' is not loaded}
    \renewcommand\color[2][]{}
  }
  \providecommand\transparent[1]{
    \errmessage{(Inkscape) Transparency is used (non-zero) for the text in Inkscape, but the package 'transparent.sty' is not loaded}
    \renewcommand\transparent[1]{}
  }
  \providecommand\rotatebox[2]{#2}
  \newcommand*\fsize{\dimexpr\f@size pt\relax}
  \newcommand*\lineheight[1]{\fontsize{\fsize}{#1\fsize}\selectfont}
  \ifx\svgwidth\undefined
    \setlength{\unitlength}{430.49870529bp}
    \ifx\svgscale\undefined
      \relax
    \else
      \setlength{\unitlength}{\unitlength * \real{\svgscale}}
    \fi
  \else
    \setlength{\unitlength}{\svgwidth}
  \fi
  \global\let\svgwidth\undefined
  \global\let\svgscale\undefined
  \makeatother
  \begin{picture}(1,0.37060145)
    \lineheight{1}
    \setlength\tabcolsep{0pt}
    \put(0,0){\includegraphics[width=\unitlength,page=1]{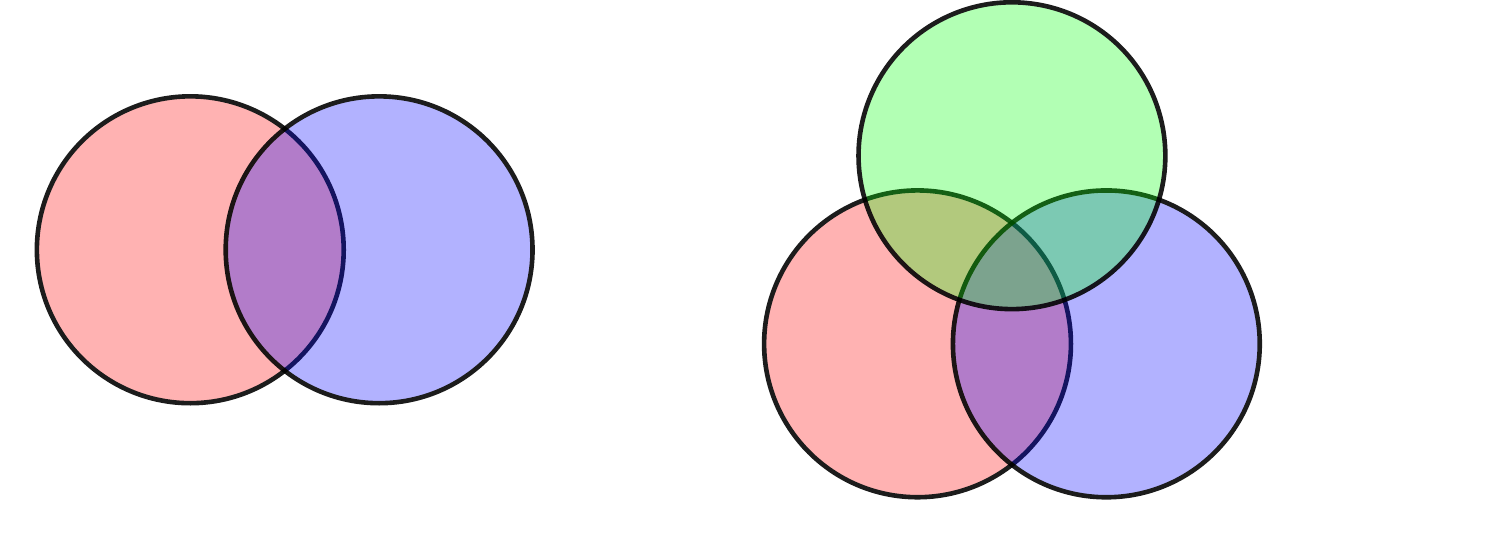}}
    \put(0.57287867,0.19333056){\color[rgb]{0,0,0}\makebox(0,0)[lt]{\begin{minipage}{0.21180496\unitlength}\centering $\Info\(\A,\B,\C\)$\end{minipage}}}
    \put(0.6643308,0.10985173){\color[rgb]{0,0,0}\makebox(0,0)[lt]{\begin{minipage}{0.21180496\unitlength}\centering $\Ent\(\B|\A,\C\)$\end{minipage}}}
    \put(0.46330959,0.12378906){\color[rgb]{0,0,0}\makebox(0,0)[lt]{\begin{minipage}{0.21180496\unitlength}\centering $\Ent\(\A|\B,\C\)$\end{minipage}}}
    \put(0.54814069,0.30488799){\color[rgb]{0,0,0}\makebox(0,0)[lt]{\begin{minipage}{0.21180496\unitlength}\centering $\Ent\(\C|\A,\B\)$\end{minipage}}}
    \put(0.62774531,0.2238773){\color[rgb]{0,0,0}\makebox(0,0)[lt]{\begin{minipage}{0.21180496\unitlength}\centering $\Info\(\B,\C|\A\)$\end{minipage}}}
    \put(0.52844191,0.23084597){\color[rgb]{0,0,0}\makebox(0,0)[lt]{\begin{minipage}{0.21180496\unitlength}\centering $\Info\(\A,\C|\B\)$\end{minipage}}}
    \put(0.58821051,0.14427267){\color[rgb]{0,0,0}\makebox(0,0)[lt]{\begin{minipage}{0.21180496\unitlength}\centering $\Info\(\A,\B|\C\)$\end{minipage}}}
    \put(0.63390612,0.03371002){\color[rgb]{0,0,0}\makebox(0,0)[lt]{\begin{minipage}{0.21180496\unitlength}\centering $\Ent\(\C\)$\end{minipage}}}
    \put(0.3581075,0.08981676){\color[rgb]{0,0,0}\makebox(0,0)[lt]{\begin{minipage}{0.21180496\unitlength}\centering $\Ent\(\A\)$\end{minipage}}}
    \put(0.43610713,0.32481025){\color[rgb]{0,0,0}\makebox(0,0)[lt]{\begin{minipage}{0.21180496\unitlength}\centering $\Ent\(\B\)$\end{minipage}}}
    \put(0.7998492,0.29880383){\color[rgb]{0,0,0}\makebox(0,0)[lt]{\begin{minipage}{0.21180496\unitlength}\centering $\VarInfo\(\A,\B,\C\)$\end{minipage}}}
    \put(0,0){\includegraphics[width=\unitlength,page=2]{VI.pdf}}
    \put(0.19836327,0.20611956){\color[rgb]{0,0,0}\makebox(0,0)[lt]{\begin{minipage}{0.21180496\unitlength}\centering $\Ent\(\B|\A\)$\end{minipage}}}
    \put(-0.0113345,0.21604487){\color[rgb]{0,0,0}\makebox(0,0)[lt]{\begin{minipage}{0.21180496\unitlength}\centering $\Ent\(\A|\B\)$\end{minipage}}}
    \put(0.08883999,0.22015579){\color[rgb]{0,0,0}\makebox(0,0)[lt]{\begin{minipage}{0.21180496\unitlength}\centering $\Info\(\A,\B\)$\end{minipage}}}
    \put(0.02140974,0.08981676){\color[rgb]{0,0,0}\makebox(0,0)[lt]{\begin{minipage}{0.21180496\unitlength}\centering $\Ent\(\A\)$\end{minipage}}}
    \put(0.14772511,0.08981676){\color[rgb]{0,0,0}\makebox(0,0)[lt]{\begin{minipage}{0.21180496\unitlength}\centering $\Ent\(\B\)$\end{minipage}}}
    \put(0.08861356,0.35419647){\color[rgb]{0,0,0}\makebox(0,0)[lt]{\begin{minipage}{0.21180496\unitlength}\centering $\VarInfo\(\A,\B\)$\end{minipage}}}
    \put(0,0){\includegraphics[width=\unitlength,page=3]{VI.pdf}}
  \end{picture}
\endgroup

	\caption{Visualization of VI as Venn digram.}
	\label{fig:VI}
\end{figure*}

\subsection{Extension to three Modalities}

It should be clear that both approaches, proposed and Suzuki's \cite{Suzuki2017}, can be extended to multiple modalities.
In the following, an example for three modalities is given.

First, the conditional probability is investigated:

\begin{align}
\p\( \a | \b,\c\) &= \frac{\p\(\a,\b,\c,z\)}{\p\( \a,\b,\c,z\)} \frac{\p\(\a,\b,\c\)}{\p\( \b,\c\)} & \optext{multiplied by 1 \& \autoref{eq:bayes_multi}}\\
&= \frac{\p\(z,\a|\b,\c\)}{\p\( z|\a,\b,\c\)}  & \optext{\autoref{eq:bayes_multi}} \\
&= \frac{1}{\p\( z|\a,\b,\c\)} \p\(z,\a|\b,\c\) & \optext{reorder} \\
&= \frac{1}{\p\( z|\a,\b,\c\)} \frac{\p\(z,\a,\b,\c\)}{\p\( \b,\c \)} & \taref{eq:bayes_multi} \\
&= \frac{1}{\p\( z|\a,\b,\c\)} \frac{\p\(\a,\b,\c|z\) \p\(z\)}{\p\( \b,\c \)} & \taref{eq:iid} \\
&= \frac{1}{\p\( z|\a,\b,\c\)} \frac{\p\(\a|z\) \p\(\b,\c|z\) \p\(z\)}{\p\( \b,\c \)} & \taref{eq:bayes_multi} \\
&= \frac{1}{\p\( z|\a,\b,\c\)} \frac{\p\(\a|z\) \p\(z|\c,\b\) \frac{\p\(\c,\b\)}{\p\(z\)} \p\(z\)}{\p\( \b,\c \)} & \taref{eq:bayes_multi} \\
&= \frac{\p\(\a|z\) \p\(z|\c,\b\)}{\p\( z|\a,\b,\c\)}  &  
\end{align}

The log-likelihood of a single joint distribution can be written as:

\begin{align}
\log\(\p\( \a | \b,\c\)\) &= \sum_z \q \( z | \a,\b,\c \)  \log \( \frac{\p \( z,\a|\b,\c \)}{ \q \( z|\a,\b,\c \) }  \)
+
\sum_z \q \( z | \a,\b,\c \)  \log \( \frac{\q \( z|\a,\b,\c \)}{ \p \( z|\a,\b,\c \) }  \)& \\
&= \elboTMa +  \dkl \( \q \( z | \a,\b,\c \) \| \p\( z| \a,\b,\c \) \)  & \label{eq:elbo3_conditional}\\
&\geq \elboTMa &
\label{eq:elbo3_conditional_geq}
\end{align}

\subsubsection{JMVAE for three Modalities}

The log-likelihood of the VI between three distributions can be written as:

\begin{align}
L_\text{3M} &= \log\(\p\( \a | \b,\c\)\) + \log\(\p\( \b | \a,\c\)\) + \log\(\p\( \c | \b,\c\)\)& \\
&= \elboTMa + \elboTMb + \elboTMc + 3 \dkl \( \q \( z | \a,\b,\c \) \| \p\( z| \a,\b,\c \) \) & \taref{eq:elbo_conditional}\\
&\geq \elboTMa + \elboTMb + \elboTMc & 
\label{eq:elbo3_joint_geq}
\end{align}

The combined ELBO can then be rewritten as

\begin{align}
\elboTMa + \elboTMb + \elboTMc & = \EX_{\q\( z | \a,\b,\c \)} \log \( \p \( \a | z \)  \) - \dkl \( \q \( z | \a,\b,\c \) \| \p\( z| \c,\b \) \) & \\
&\phantom{=} +
\EX_{\q\( z | \a,\b,\c \)} \log \( \p \( \b | z \)  \) - \dkl \( \q \( z | \a,\b,\c \) \| \p\( z| \a,\c \) \)& \\
&\phantom{=} +
\EX_{\q\( z | \a,\b,\c \)} \log \( \p \( \c | z \)  \) - \dkl \( \q \( z | \a,\b,\c \) \| \p\( z| \a,\b \) \)& \optext{\autoref{eq:kld}}\\
&\geq \elboTJ -\dkl \( \q \( z | \a,\b,\c \) \| \p\( z| \b,\c \) \) & \\
&\phantom{=} - \dkl \( \q \( z | \a,\b,\c \) \| \p\( z| \a,\c \) \) - \dkl \( \q \( z | \a,\b,\c \) \| \p\( z| \b,\c \) \)& \label{eq:multi_vae_loss}\\
&:= \elboTM &
\end{align}

$\elboTJ$ is the joint ELBO of a joint probability distribution having three arguments (i.e. $\a$, $\b$, $\c$).
The derivation is analog to \autoref{sec:joint_VAE}.
The next steps are the application of encoders and decoders for this network which is straight forward and should be clear to the reader.

However, if we investigate the last equations, the following properties can be identified:
There are the common reconstruction terms ($\EX$) for each decoder $\p\(*|z \)$ wrt. the full multi-modal decoder $\q\( z|\a,\b,\c \)$.
The KL terms show the \textbf{drawback} of the VI approach.
As before, these regularizer tend to bring the encoders' distribution to match each other.
But only pairwise encoders (e.g. $\p\( z|\a,\b \)$) remain and thus, uni-modal encoders are neglected.

This means from a practical point of view, that when we have $N$ modalities in a setup, we only can build derived setups having $N-1$ modalities.

\subsubsection{Proposed JMVAE for three Modalities}

The derivation from the joint log likelihood can be written analogously:

\begin{align}
\log\(\p\( \a,\b,\c \) \) &= \frac{3}{3} \log\(\p\( \a,\b,\c \) \) & \\
&= \frac{1}{3} \log\(\p\( \a,\b,\c \)^3 \) & \\
&= \frac{1}{3} \log\(\p\( \a,\b,\c \)\p\( \a,\b,\c \)\p\( \a,\b,\c \) \) & \\
&= \frac{1}{3} \log\(\p\(\a,\b \)\p\(\b,\c \)\p\(\a,\c \)\p\( \a|\b,\c \)\p\( \b|\a,\c \)\p\( \c|\a,\b \) \) & \\
&= \frac{1}{3} \(\log\(\p\(\a,\b \) \)+\log\( \p\(\b,\c \) \)+\log\( \p\(\a,\c \) \)+\log\( \p\( \a|\b,\c \) \)+\log\( \p\( \b|\a,\c \) \)+\log\( \p\( \c|\a,\b \) \)\) & \\
&= \frac{1}{3} \(\frac{2}{2}\(\log\(\p\(\a,\b \) \)+\log\( \p\(\b,\c \) \)+\log\( \p\(\a,\c \) \)\)+\log\( \p\( \a|\b,\c \) \)+\log\( \p\( \b|\a,\c \) \)+\log\( \p\( \c|\a,\b \) \)\) & \\
&= \frac{1}{6}\(\log\(\p\(\a,\b \)^2 \)+\log\( \p\(\b,\c \)^2 \)+\log\( \p\(\a,\c \)^2 \)\) & \\
&\phantom{=} + \frac{1}{3} \(\log\( \p\( \a|\b,\c \) \)+\log\( \p\( \b|\a,\c \) \)+\log\( \p\( \c|\a,\b \) \)\) & \\
&= \frac{1}{6}\(\log\(\p\(a\) \p\(b\)\p\(\a|\b \)p\(\b|\a \) \)+\log\( \p\(c\) \p\(b\)\p\(\c|\b \)p\(\b|\c \) \)+\log\( \p\(a\) \p\(c\)\p\(\a|\c \)p\(\c|\a \) \)\) & \\
&\phantom{=} + \frac{1}{3} \(\log\( \p\( \a|\b,\c \) \)+\log\( \p\( \b|\a,\c \) \)+\log\( \p\( \c|\a,\b \) \)\) & \\
&= \frac{1}{6}\(\log\(\p\(\a|\b \)\) + \log\(p\(\b|\a \) \)+\log\(\p\(\c|\b \)\) + \log\(p\(\b|\c \) \)+\log\( \p\(\a|\c \)\) + \log\(p\(\c|\a \) \)\) & \\
&\phantom{=} + \frac{1}{3} \(\log\( \p\(a\)\) + \log\( \p\(b\)\) + \log\( \p\(c\)\) + \log\( \p\( \a|\b,\c \) \)+\log\( \p\( \b|\a,\c \) \)+\log\( \p\( \c|\a,\b \) \)\) &   
\end{align}

It is now straight forward, by applying all former mentioned equations, to derive the ELBO for the above marginal log-likelihood.
As one can imagine, the above equation results in a pretty heavy loss term but with the big advantage of respecting all permutations of modalities.

This means again from a practical point of view, in comparison to the approach by Suzuki et al. \cite{Suzuki2017}, that we can build arbitrary sensor setups having $1$ to $N$ modalities.

\bibliographystyle{IEEEtran}
\bibliography{main}

\begin{thebibliography}{1}
\providecommand{\url}[1]{#1}
\csname url@rmstyle\endcsname
\providecommand{\newblock}{\relax}
\providecommand{\bibinfo}[2]{#2}
\providecommand\BIBentrySTDinterwordspacing{\spaceskip=0pt\relax}
\providecommand\BIBentryALTinterwordstretchfactor{4}
\providecommand\BIBentryALTinterwordspacing{\spaceskip=\fontdimen2\font plus
\BIBentryALTinterwordstretchfactor\fontdimen3\font minus
  \fontdimen4\font\relax}
\providecommand\BIBforeignlanguage[2]{{%
\expandafter\ifx\csname l@#1\endcsname\relax
\typeout{** WARNING: IEEEtran.bst: No hyphenation pattern has been}%
\typeout{** loaded for the language `#1'. Using the pattern for}%
\typeout{** the default language instead.}%
\else
\language=\csname l@#1\endcsname
\fi
#2}}

\bibitem{DBLP:journals/corr/KingmaW13}
\BIBentryALTinterwordspacing
D.~P. Kingma and M.~Welling, ``{Auto-Encoding Variational Bayes},''
  \emph{CoRR}, vol. abs/1312.6, 2013.
\BIBentrySTDinterwordspacing

\bibitem{Suzuki2017}
M.~Suzuki, K.~Nakayama, and Y.~Matsuo, ``{Joint multimodal learning with deep
  generative models},'' pp. 1--12, 2017.

\end{thebibliography}

\end{document}